\documentclass{article}
\usepackage{ijcai17}

\usepackage{times}

\usepackage{graphicx}
\usepackage{subfigure}
\usepackage{url}
\usepackage{latexsym}
\usepackage{bm}
\usepackage{amsmath}
\usepackage{amssymb}
\usepackage{amsthm}
\usepackage{amsfonts}
\usepackage{algorithm}
\usepackage{algorithmic}
\usepackage{epstopdf}
\usepackage{float}
\usepackage{caption}
\usepackage{multirow}
\title{Learning of Human-like Algebraic Reasoning Using \\ Deep Feedforward Neural Networks}
\author{
Cheng-Hao Cai$^1$ ~~ Dengfeng Ke$^1$\thanks{Corresponding Author.} ~~ Yanyan Xu$^2$ ~~ Kaile Su$^3$\\
$^1$National Laboratory of Pattern Recognition, Institute of Automation, Chinese Academy of Sciences\\
$^2$School of Information Science and Technology, Beijing Forestry University\\
$^3$Institute for Integrated and Intelligent Systems, Griffith University\\
chenghao.cai@outlook.com, dengfeng.ke@nlpr.ia.ac.cn, xuyanyan@bjfu.edu.cn, k.su@griffith.edu.au
}

\begin{document}

\maketitle

\begin{abstract}
  There is a wide gap between symbolic reasoning and deep learning. In this research, we explore the possibility of using deep learning to improve symbolic reasoning. Briefly, in a reasoning system, a deep feedforward neural network is used to guide rewriting processes after learning from algebraic reasoning examples produced by humans. To enable the neural network to recognise patterns of algebraic expressions with non-deterministic sizes, reduced partial trees are used to represent the expressions. Also, to represent both top-down and bottom-up information of the expressions, a centralisation technique is used to improve the reduced partial trees. Besides, symbolic association vectors and rule application records are used to improve the rewriting processes. Experimental results reveal that the algebraic reasoning examples can be accurately learnt only if the feedforward neural network has enough hidden layers. Also, the centralisation technique, the symbolic association vectors and the rule application records can reduce error rates of reasoning. In particular, the above approaches have led to 4.6\% error rate of reasoning on a dataset of linear equations, differentials and integrals.
\end{abstract}

\section{Introduction}

It is challenging to integrate symbolic reasoning and deep learning in effective ways \cite{Garcez:Neural}. In the field of symbolic reasoning, much work has been done on using formal methods to model reliable reasoning processes \cite{DBLP:books/daglib/0070484}. For instance, algebraic reasoning can be modelled by using first-order predicate logics or even higher-order logics, but these logics are usually designed by experienced experts, because it is challenging for machines to learn these logics from data automatically \cite{DBLP:journals/ai/BundyW81,DBLP:books/sp/NipkowPW02}. On the other hand, recent approaches on deep learning have revealed that deep neural networks are powerful tools for learning from data \cite{Lecun2015Deep}, especially for learning speech features \cite{DBLP:journals/taslp/MohamedDH12} and image features \cite{DBLP:journals/corr/SunLWT15}. However, not much work has been done on using deep neural networks to learn formal symbolic logics. To close the gap between symbolic reasoning and deep learning, this research explores the possibility of using deep feedforward neural networks to learn logics of rewriting in algebraic reasoning. In other words, we try to teach neural networks to solve mathematical problems, such as finding the solution of an equation and calculating the differential or integral of an expression, by using a rewriting system.

Rewriting is an important technique in symbolic reasoning. Its core concept is to simply reasoning process by using equivalence relations between different expressions \cite{Bundy1983The}. Usually, rewriting is based on a tree-manipulating system, as many algebraic expressions can be represented by using tree structures, and the manipulation of symbols in the expressions is equivalent to the manipulation of nodes, leaves and sub-trees on the trees \cite{Rosen1973Tree}. To manipulate symbols, a rewriting system usually uses one way matching, which is a restricted application of unification, to find a desired pattern from an expression and then replaces the pattern with another equivalent pattern \cite{Bundy1983The}. In order to reduce the search space, rewriting systems are expected to be Church-Rosser, which means that they should be terminating and locally confluent \cite{Rosen1973Tree,Huet1980Confluent}. Thus, very careful designs and analyses are needed: A design can start from small systems, because proving termination and local confluence of a smaller system is usually easier than proving those of a larger system \cite{DBLP:journals/ai/BundyW81}. Some previous work has focused on this aspect: The Knuth-Bendix completion algorithm can be used to solve the problem of local confluence \cite{Knuth1983Simple}, and Huet \shortcite{DBLP:journals/jcss/Huet81} has provided a proof of correctness for this algorithm. Also, dependency pairs \cite{DBLP:journals/tcs/ArtsG00} and semantic labelling \cite{DBLP:journals/fuin/Zantema95} can solve the problem of termination for some systems. After multiple small systems have been designed, they can be combined into a whole system, because the direct sum of two Church-Rosser systems holds the same property \cite{DBLP:journals/jacm/Toyama87}.

Deep neural networks have been used in many fields of artificial intelligence, including speech recognition \cite{DBLP:journals/taslp/MohamedDH12}, human face recognition \cite{DBLP:journals/corr/SunLWT15}, natural language understanding \cite{DBLP:journals/taslp/SarikayaHD14}, reinforcement learning for playing video games \cite{mnih2015human} and Monte Carlo tree search for playing Go \cite{Silver_2016}. Recently, some researchers are trying to extend them to reasoning tasks. For instance, Irving et al. \shortcite{DBLP:conf/nips/IrvingSAECU16} have proposed DeepMath for automated theorem proving with deep neural networks. Also, Serafini and Garcez \shortcite{DBLP:journals/corr/SerafiniG16} have proposed logic tensor networks to combine deep learning with logical reasoning. In addition, Garnelo et al. \shortcite{DBLP:journals/corr/GarneloAS16} have explored deep symbolic reinforcement learning.

In this research, we use deep feedforward neural networks \cite{Lecun2015Deep} to guide rewriting processes. This technique is called human-like rewriting, as it is adapted from standard rewriting and can simulate human's behaviours of using rewrite rules after learning from algebraic reasoning schemes. The following sections provide detailed discussions about this technique: Section \ref{hlrrt} introduces the core method of human-like rewriting. Section \ref{ars} discusses algebraic reasoning schemes briefly. Section \ref{optimi} provides three methods for system improvement. Section \ref{experi} provides experimental results of the core method and the improvement methods. Section \ref{conclus} is for conclusions.

\section{Human-like Rewriting}
\label{hlrrt}

Rewriting is an inference technique for replacing expressions or subexpressions with equivalent ones \cite{Bundy1983The}. For instance\footnote{We use the mathematical convention that a word is a constant if its first letter is in upper case, and it is a variable if its first letter is in lower case.}, given two rules of the Peano axioms:
\begin{equation}
\label{peano1}
x + 0 \Rightarrow x
\end{equation}
\begin{equation}
\label{peano2}
x + S(y) \Rightarrow S(x+y)
\end{equation}
$ S(0)+S(S(0)) $ can be rewritten via:
\begin{equation}
\begin{aligned}
&~~\underbrace{S(0)+S(S(0))}_{by~(\ref{peano2})} \\
\Rightarrow & ~~S(~\underbrace{S(0)+S(0)}_{by~(\ref{peano2})}~) \\
\Rightarrow & ~~S(S(~\underbrace{S(0)+0}_{by~(\ref{peano1})}~)) \\
\Rightarrow & ~~S(S(S(0)))
\end{aligned}
\end{equation}
More detailed discussions about the Peano axioms can be found from \cite{Pillay1981Models}. Generally, rewriting requires a source expression $ s $ and a set of rewrite rules $ \tau $. Let $ l \Rightarrow r $ denote a rewrite rule in $ \tau $, $ t $ a subexpression of $ s $, and $ \theta $ the most general unifier of one way matching from $ l $ and $ t $. A single rewriting step of inference can be formed as:
\begin{equation}
\label{rewriteinf}
\cfrac{s(t)~~~~~(l \Rightarrow r) \in \tau~~~~~l[\theta] \equiv t}{s(r[\theta])}
\end{equation}
It is noticeable that $ \theta $ is only applied to $ l $, but not to $ t $. The reason is that one way matching, which is a restricted application of unification, requires that all substitutions in a unifier are only applied to the left-hand side of a unification pair. Standard rewriting is to repeat the above step until no rule can be applied to the expression further. It requires the set of rewrite rules $ \tau $ to be Church-Rosser, which means that $ \tau $ should be terminating and locally confluent. This requirement restricts the application of rewriting in many fields. For instance, the chain rule in calculus $ \cfrac{D(f)}{D(x)} \Rightarrow \cfrac{D(f)}{D(u)} \cdot \cfrac{D(u)}{D(x)} $, which is very important for computing derivatives, will result in non-termination:
\begin{equation}
\begin{aligned}
&~~ \cfrac{D(Sin(X))}{D(X)} \\
\Rightarrow & ~~ \cfrac{D(Sin(X))}{D(u_1)} \cdot \cfrac{D(u_1)}{D(X)} \\
\Rightarrow & ~~ \cfrac{D(Sin(X))}{D(u_2)} \cdot \cfrac{D(u_2)}{D(u_1)} \cdot \cfrac{D(u_1)}{D(X)} \\
\Rightarrow & ~~ \cfrac{D(Sin(X))}{D(u_3)} \cdot \cfrac{D(u_3)}{D(u_2)} \cdot \cfrac{D(u_2)}{D(u_1)} \cdot \cfrac{D(u_1)}{D(X)} \\
\Rightarrow & ~~ \cdots
\end{aligned}
\end{equation}
The above process means that it is challenging to use the chain rule in standard rewriting. Similarly, a commutativity rule $ x~\circ~y \Rightarrow y~ \circ~x $, where $ \circ $ is an addition, a multiplication, a logical conjunction, a logical disjunction or another binary operation satisfying commutativity, is difficult to be used in standard rewriting. If termination is not guaranteed, it will be difficult to check local confluence, as local confluence requires a completely developed search tree, but non-termination means that the search tree is infinite and cannot be completely developed. More detailed discussion about standard rewriting and Church-Rosser can be found from \cite{Bundy1983The}.

Human-like rewriting is adapted from standard rewriting. It uses a deep feedforward neural network \cite{Lecun2015Deep} to guide rewriting processes. The neural network has learnt from some rewriting examples produced by humans, so that it can, to some extent, simulate human's ways of using rewrite rules: Firstly, non-terminating rules are used to rewrite expressions. Secondly, local confluence is not checked. Lastly, experiences of rewriting can be learnt and can guide future rewriting processes.

\begin{figure}[!htb]
\centering
\includegraphics[width=6cm]{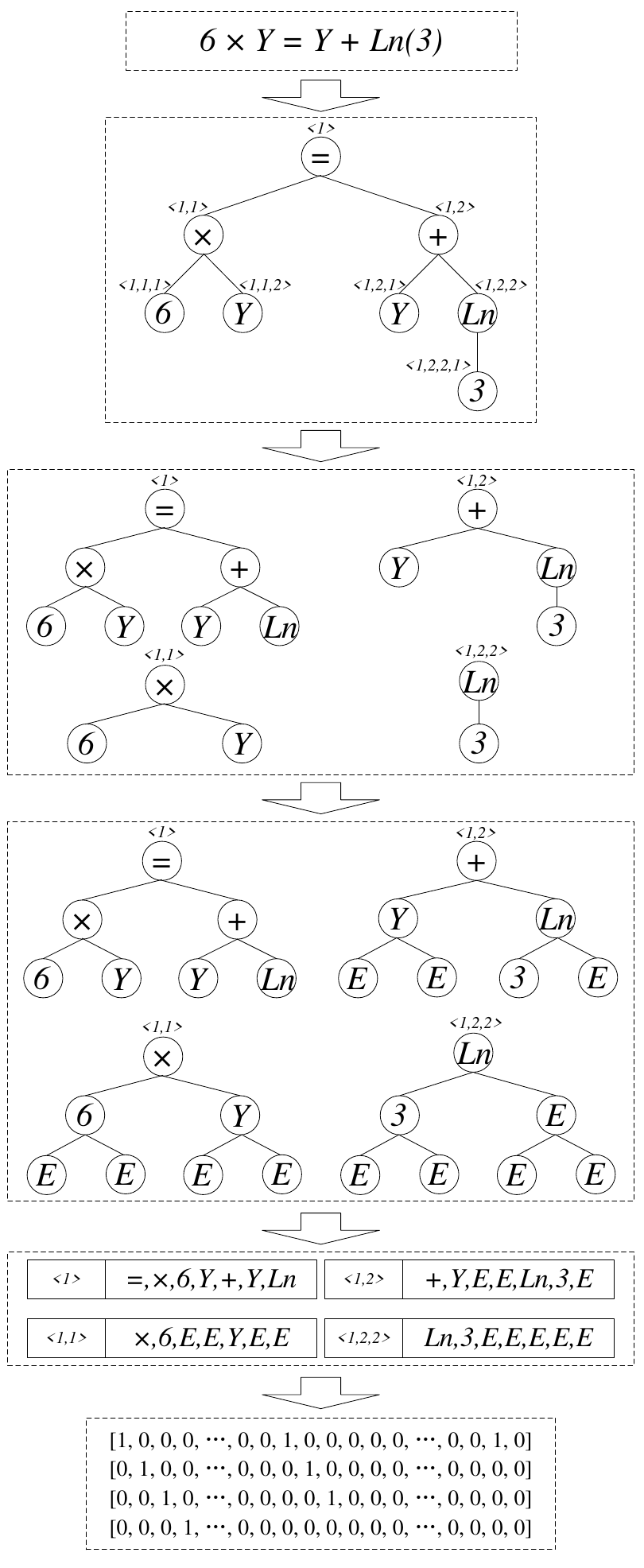}
\caption{An expression $ 6 \times Y = Y + Ln(3) $ is transformed to the RPT representation. The maximum depth and the breadth of a tree are defined as 2. ``$E$" is an abbreviation of ``Empty".}
\label{fig:rpt}
\end{figure}

To train the feedforward neural network, input data and target data are required. An input can be generated via the following steps: Firstly, an expression is transformed to a parsing tree \cite{DBLP:books/daglib/0017977} with position annotations. A position annotation is a unique label $ <p_1,p_2,\cdots,p_N> $ indicating a position on a tree, where each $p_i$ is the order of a branch. Then the tree is reduced to a set of partial trees with a predefined maximum depth $ d $. Next, the partial trees are expanded to perfect $k$-ary trees with the depth $ d $ and a predefined breadth $ k $. In particular, empty positions on the prefect $k$-ary trees are filled by $ Empty $. After that, the perfect $k$-ary trees are transformed to lists via in-order traversal. Detailed discussions about perfect $k$-ary trees and in-order traversal can be found from \cite{Cormen2001Introduction}. Finally, the lists with their position annotations are transformed to a set of one-hot representations \cite{DBLP:conf/acl/TurianRB10}. In particular, $ Empty $ is transformed to a zero block. Figure \ref{fig:rpt} provides an example for the above procedure. This representation is called a reduced partial tree (RPT) representation of the expression. A target is the one-hot representation \cite{DBLP:conf/acl/TurianRB10} of a rewrite rule name with a position annotation for applying the rule.

It is noticeable that the input of the neural network is a set of vectors, and the number of vectors is non-deterministic, as it depends on the structure of the expression. However, the target is a single vector. Thus, the dimension of the input will disagree with the dimension of the target if a conventional feedforward neural network structure is used. To solve this problem, we replace its Softmax layer with an averaged Softmax layer. Let $ x_{j,i} $ denote the $i$th element of the $j$th input vector, $P$ the number of the input vectors, $ \bm{u} $ an averaged input vector, $u_i$ the $i$th element of $\bm{u}$, $\bm{W}$ a weight matrix, $\bm{b}$ a bias vector, $Softmax$ the standard Softmax function \cite{Bishop2006Pattern}, and $\bm{y}$ the output vector. The averaged Softmax layer is defined as:
\begin{equation}
u_i = \cfrac{1}{P}\sum \limits_{j=1}^{P} x_{j,i}
\end{equation}
\begin{equation}
\bm{y} = Softmax(\bm{W} \cdot \bm{u} + \bm{b})
\end{equation}
It is noticeable that the output is a single vector regardless of the number of the input vectors.

The feedforward neural network is trained by using the back-propagation algorithm with the cross-entropy error function \cite{DBLP:journals/nn/Hecht-Nielsen88a,Bishop2006Pattern}. After training, the neural network can be used to guide a rewriting procedure: Given the RPT representation of an expression, the neural network uses forward computation to get an output vector, and the position of the maximum element indicates the name of a rewrite rule and a possible position for the application of the rule.

\section{Algebraic Reasoning Schemes}
\label{ars}

The learning of the neural network is based on a set of algebraic reasoning schemes. Generally, an algebraic reasoning scheme consists of a question, an answer and some intermediate reasoning steps. The question is an expression indicating the starting point of reasoning. The answer is an expression indicating the goal of reasoning. Each intermediate reasoning step is a record consisting of:
\begin{itemize}
\item A source expression;
\item The name of a rewrite rule;
\item A position annotation for applying the rewrite rule;
\item A target expression.
\end{itemize}In particular, the source expression of the first reasoning step is the question, and the target expression of the final reasoning step is the answer. Also, for each reasoning step, the target expression will be the source expression of the next step if the ``next step" exists. By applying all intermediate reasoning steps, the question can be rewritten to the answer deterministically.

In this research, algebraic reasoning schemes are developed via a rewriting system in SWI-Prolog \cite{DBLP:journals/tplp/WielemakerSTL12}. The rewriting system is based on Rule (\ref{rewriteinf}), and it uses breadth-first search to find intermediate reasoning steps from a question to an answer. Like most rewriting systems and automated theorem proving systems\footnote{A practical example is the ``by auto" function of Isabelle/HOL \cite{DBLP:books/sp/NipkowPW02}. It is often difficult to prove a complex theorem automatically, so that experts' guidance is often required.}, its ability of reasoning is restricted by the problem of combinatorial explosion: The number of possible ways of reasoning can grow rapidly when the question becomes more complex \cite{Bundy1983The}. Therefore, a full algebraic reasoning scheme of a complex question is usually difficult to be generated automatically, and guidance from humans is required. In other words, if the system fails to develop the scheme, we will apply rewrite rules manually until the remaining part of the scheme can be developed automatically, or we will provide some subgoals for the system to reduce the search space. After algebraic reasoning schemes are developed, their intermediate reasoning steps are used to train the neural network: For each step, the RPT representation of the source expression is the input of the neural network, and the one-hot representation of the rewrite rule name and the position annotation is the target of the neural network, as discussed by Section \ref{hlrrt}.

\section{Methods for System Improvement}
\label{optimi}

\subsection{Centralised RPT Representation}

The RPT representation discussed before is a top-down representation of an expression: A functor in the expression is a node, and arguments dominated by the functor are child nodes or leaves of the node. However, it does not record bottom-up information about the expression. For instance, in Figure \ref{fig:rpt}, the partial tree labelled $ <1,1> $ does not record any information about its parent node ``$ = $''.

\begin{figure}[!htb]
\centering
\includegraphics[width=5.2cm]{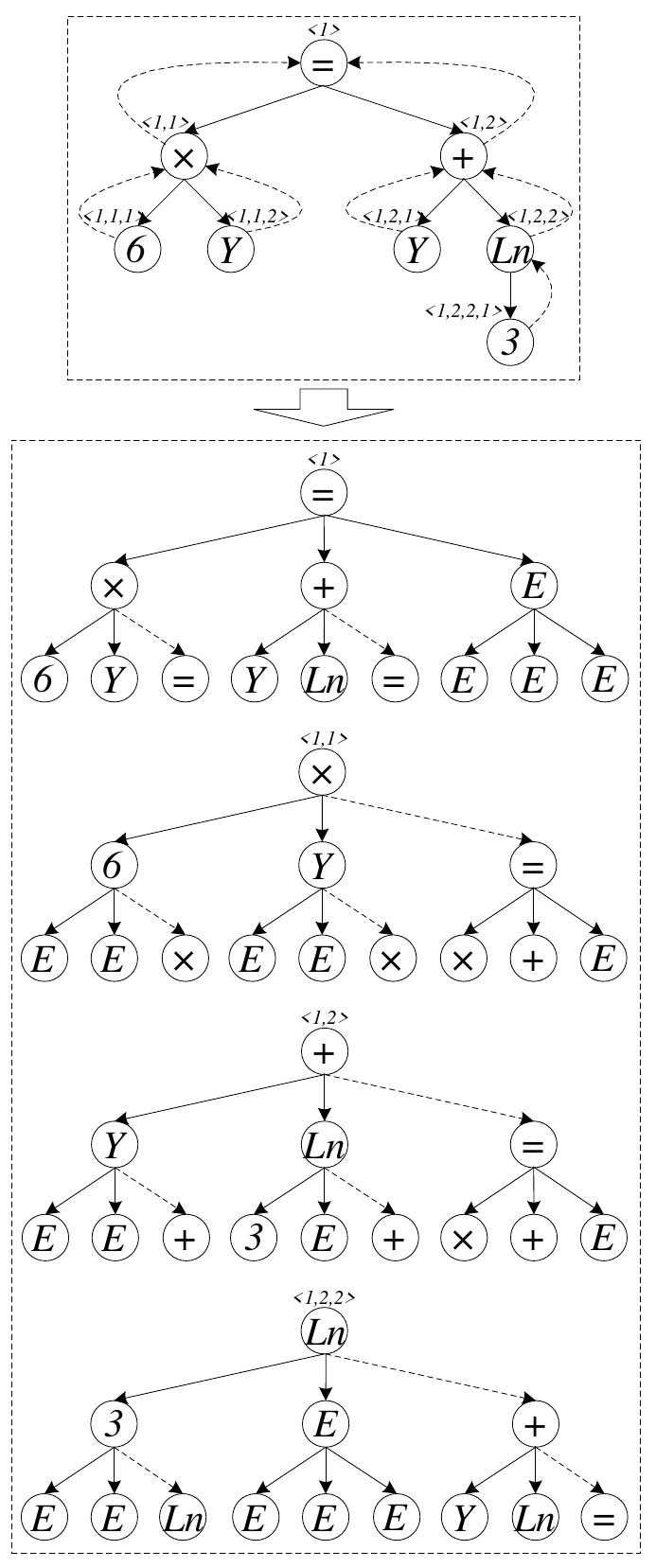}
\caption{The parsing tree of $ 6 \times Y = Y + Ln(3) $ is centralised, reduced and expanded to a set of perfect $k$-ary trees. Dashed arrows denote additional branchs generated by centralisation. 
}
\label{fig:crpt}
\end{figure}

A centralised RPT (C-RPT) representation can represent both top-down and bottom-up information of an expression: Firstly, every node on a tree considers itself as the centre of the tree and grows an additional branch to its parent node (if it exists), so that the tree becomes a directed graph. This step is called ``centralisation". Then the graph is reduced to a set of partial trees and expanded to a set of perfect $k$-ary trees. In particular, each additional branch is defined as the $ k $th branch of its parent node, and all empty positions dominated by the parent node are filled by $ Empty $. Detailed discussions about perfect $k$-ary trees and directed graphs can be found from \cite{Cormen2001Introduction}. Figure \ref{fig:crpt} provides an example for the above steps. Finally, these perfect $k$-ary trees are transformed to lists and further represented as a set of vectors, as discussed by Section \ref{hlrrt}.

\subsection{Symbolic Association Vector}

Consider the following rewrite rule:
\begin{equation}
x \times x \Rightarrow x ^ 2
\end{equation}
The application of this rule requires that two arguments of ``$ \times $" are the same. If this pattern exists in an expression, it will be a useful hint for selecting rules. In such case, the use of a symbolic association vector (SAV) can provide useful information for the neural network: Assume that $ H $ is the list representation of a perfect $k$-ary tree (which has been discussed by Section \ref{hlrrt}) with a length $ L $. $ S $ is defined as an $ L \times L $ matrix which satisfies:
\begin{equation}
S_{i,j} =
\begin{cases}
1, & ~~~\mbox{if $H_i = H_j$ and $ i \not = j $}; \\
0, & ~~~\mbox{otherwise}.
\end{cases}
\end{equation}
After the matrix is produced, it can be reshaped to a vector and be a part of an input vector of the neural network.

\subsection{Rule Application Record}

Previous applications of rewrite rules can provide hints for current and future applications. In this research, we use rule application records (RAR) to record the previous applications of rewrite rules: Let $ Q_i $ denote the $ i $th element of an RAR $ Q $, $ rule_i $ the name of the previous $i$th rewrite rule, and $ pos_i $ the position annotation for applying the rule. $ Q_i $ is defined as:
\begin{equation}
Q_i \equiv~ <rule_i,pos_i>
\end{equation}
Usually, the RAR only records the last $ N $ applications of rewrite rules, where $ N $ is a predefined length of $ Q $. To enable the neural network to read the RAR, it needs to be transformed to a one-hot representation \cite{DBLP:conf/acl/TurianRB10}. A drawback of RARs is that they cannot be used in the first $ N $ steps of rewriting, as they record exactly $ N $ previous applications of rewrite rules.

\section{Experiments}
\label{experi}

\subsection{Datasets and Evaluation Metrics}

A dataset of algebraic reasoning schemes is used to train and test models. This dataset contains 400 schemes about linear equations, differentials and integrals and 80 rewrite rules, and these schemes consist of 6,067 intermediate reasoning steps totally. We shuffle the intermediate steps and then divide them into a training set and a test set randomly: The training set contains 5,067 examples, and the test set contains 1,000 examples. After training a model with the training set, an error rate of reasoning on the test set is used to evaluate the model, and it can be computed by:
\begin{equation}
Error~Rate~=~\cfrac{N_{Error}}{N_{Total}} \times 100\%
\end{equation}
where $ N_{Error} $ is the number of cases when the model fails to indicate an expected application of rewrite rules, and $ N_{Total} $ is the number of examples in the test set.

\subsection{Using RPT Representations and Neural Networks}
\label{sec:exp1}

In this part, we evaluate the core method of human-like rewriting: All expressions in the dataset are represented by using the RPT representations. The breadth of an RPT is set to 2, because the expressions in the dataset are unary or binary. The depth of an RPT is set to 1, 2 or 3. Also, feedforward neural networks \cite{Lecun2015Deep} with 1, 3 and 5 hidden layers are used to learn from the dataset. The number of units in each hidden layer is set to 1,024, and their activation functions are rectified linear units (ReLU) \cite{DBLP:journals/jmlr/GlorotBB11}. The output layer of each neural network is an averaged Softmax layer. The neural networks are trained via the back-propagation algorithm with the cross-entropy error function \cite{DBLP:journals/nn/Hecht-Nielsen88a,Bishop2006Pattern}. When training models, learning rates are decided by the Newbob+/Train strategy \cite{DBLP:conf/icassp/WieslerRSN14}: The initial learning rate is set to 0.01, and the learning rate is halved when the average improvement of the cross-entropy loss on the training set is smaller than 0.1. The training process stops when the improvement is smaller than 0.01.

\begin{figure}[!htb]
\centering
\includegraphics[width=6cm]{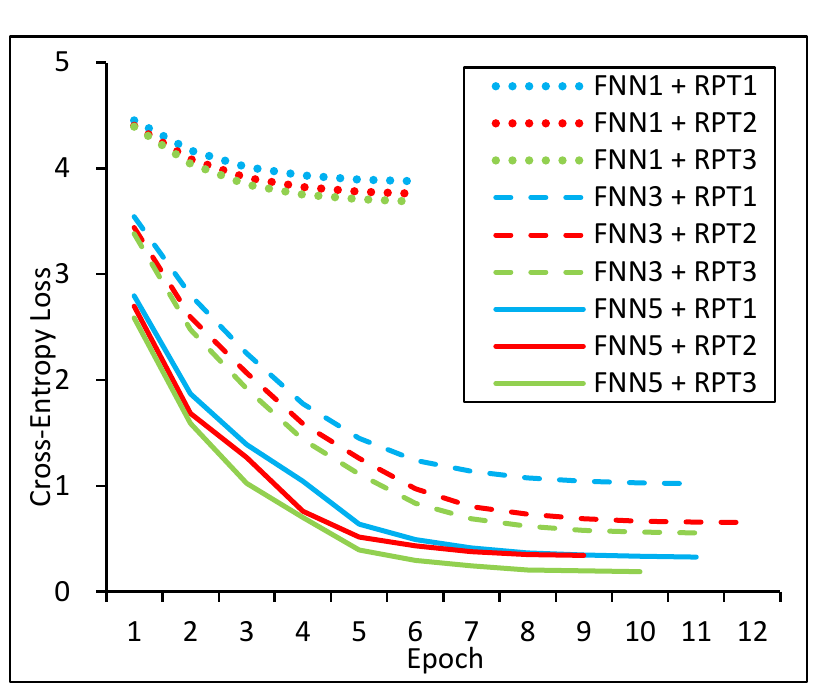}
\caption{Learning Curves of FNN + RPT Models.}
\label{fig:layers_exp}
\end{figure}

Figure \ref{fig:layers_exp} provides learning curves of the models, where ``FNN$ n $" means that the neural network has $ n $ hidden layers, and ``RPT$ m $" means that the depth of RPTs is $ m $. To aid the readability, the curves of ``FNN1", ``FNN3" and ``FNN5" are in blue, red and green respectively, and the curves of ``RPT1", ``RPT2" and ``RPT3" are displayed by using dotted lines, dashed lines and solid lines respectively. By comparing the curves with the same colour, it is noticeable that more hidden layers can bring about significantly better performance of learning. On the other hand, if the neural network only has a single hidden layer, the learning will stop early, while the cross-entropy loss is very high. Also, by comparing the curves with the same type of line, it is noticeable that a deeper RPT often brings about better performance of learning, but an exception is the curve of the ``FNN5 + RPT2" model.

\begin{table}[!htb]
\caption{Error Rates (\%) of Reasoning on the Test Set.}
\label{err_res_1}
\centering
\begin{tabular}{|c|c|c|c|}
\hline
\multirow{2}{*}{FNN$n$} & \multicolumn{3}{|c|}{RPT$m$} \\
\cline{2-4}
  & $m=1$ & $m=2$ & $m=3$ \\
\hline
$n=1$ & 80.4 & 79.9 & 76.1 \\
\hline
$n=3$ & 27.4 & 20.4 & 18.8 \\
\hline
$n=5$ & 16.5 & 16.6 & $\bm{12.9}$ \\
\hline
\end{tabular}
\end{table}

Table \ref{err_res_1} reveals performance of the trained models on the test set. In this table, results in ``FNN$n$" rows and ``RPT$m$" columns correspond to the ``FNN$n$+RPT$m$" models in Figure \ref{fig:layers_exp}. It is noticeable that the error rates of reasoning decrease significantly when the numbers of hidden layers increase. Also, the error rates of reasoning often decrease when the depths of RPTs increase, but an exception occurs in the case of ``FNN5 + RPT2". We believe that the reason why the exception occurs is that the learning rate strategy results in early stop of training. In addition, the error rate of the FNN5 + RPT3 model is the best among all results.

\subsection{Using Improvement Methods}

In Section \ref{sec:exp1}, we have found that the neural networks with 5 hidden layers have better performance than those with 1 or 3 hidden layers on the task of human-like rewriting. Based on the neural networks with 5 hidden layers, we apply the three improvement methods to these models.

\begin{figure*}[!htb]
\centering
\subfigure[C-RPT Models.]{ \label{fig:crpt_exp} 
\includegraphics[width=6cm]{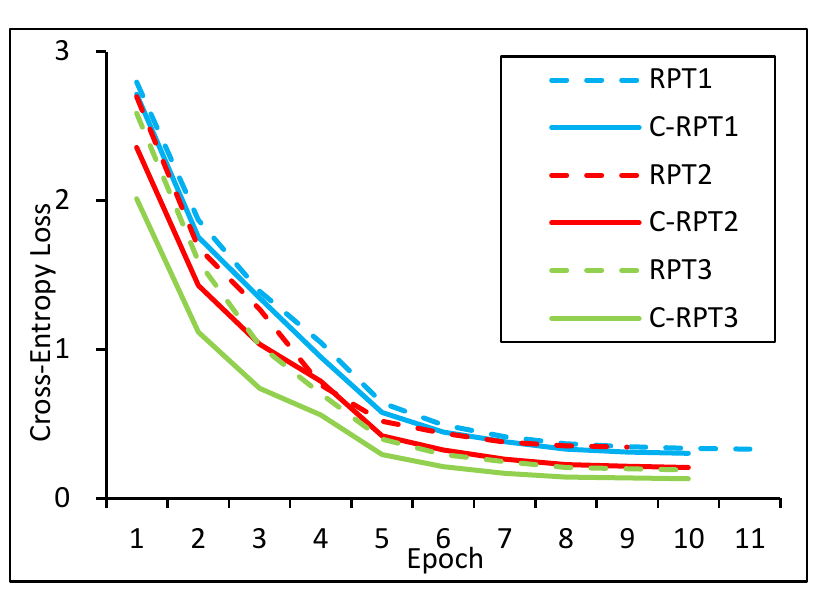}
}
\subfigure[SAV Models.]{
\label{fig:sav_exp} 
\includegraphics[width=6cm]{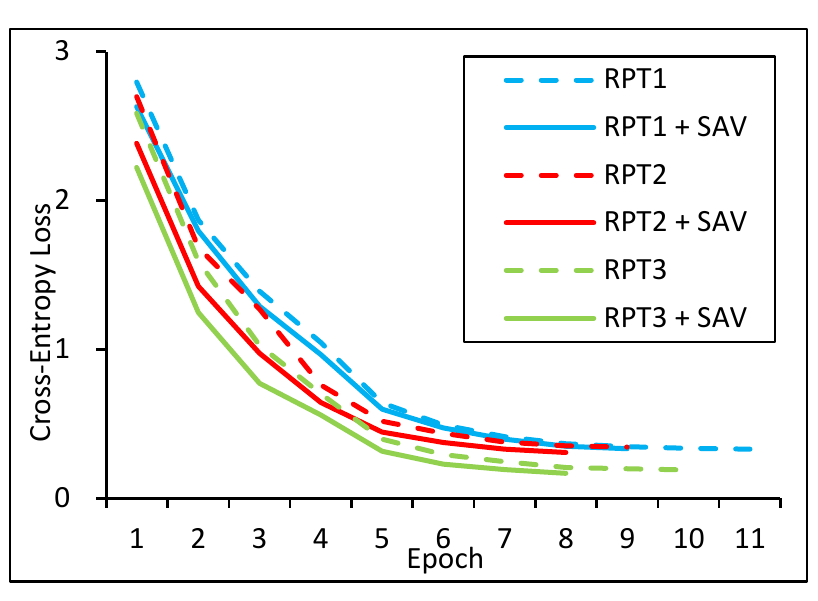}
}
\subfigure[RAR Models.]{
\label{fig:rr_exp} 
\includegraphics[width=6cm]{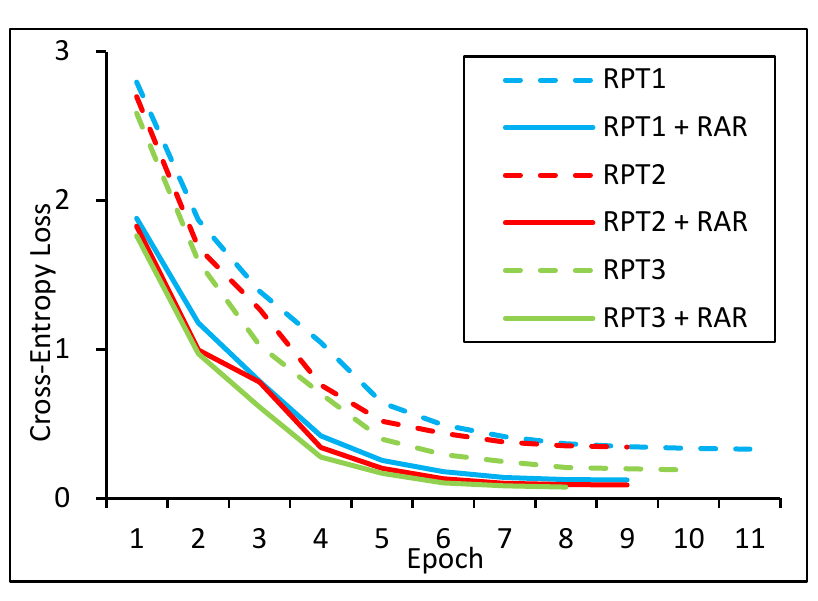}
}
\subfigure[C-RPT + SAV + RAR Models.]{
\label{fig:crpt_sav_rr_exp} 
\includegraphics[width=6cm]{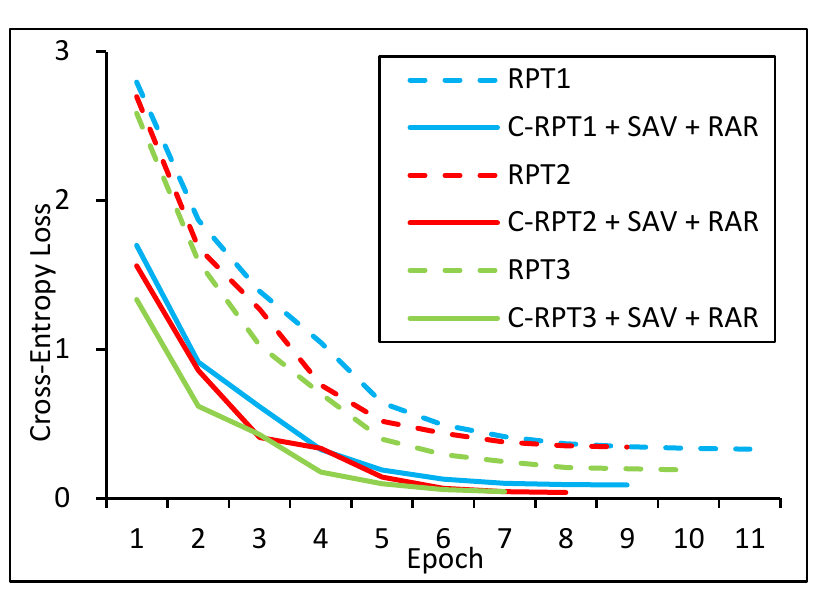}
}
\caption{Learning Curves of Models Improved by C-RPTs, SAVs and RARs.}
\label{fig:opt} 
\end{figure*}

Figure \ref{fig:opt} shows learning curves of models improved by C-RPTs, SAVs and RARs. Also, learning curves of the baseline RPT$m$ models are displayed by using dashed lines, where $ m $ is the depth of RPTs. Learning curves of the C-RPT models are displayed by Figure \ref{fig:crpt_exp}. A comparison between two lines in the same colour reveals that the C-RPT representation can improve the model when $m$ is fixed. Also, the C-RPT2 curve is very close to the RPT3 curve during the last 6 epochs, which reveals that there might be a trade-off between using C-RPTs and increasing the depth of RPTs. The best learning curve is the C-RPT3 curve, as its cross-entropy loss is always the lowest during all epochs. Figure \ref{fig:sav_exp} provides learning curves of the RPT models with the SAV method. It is noticeable that SAVs have two effects: The first is that they can bring about lower cross-entropy losses. The second is that they can reduce the costs of learning time, as each RPT$m$ + SAV model uses fewer epochs to finish learning than its counterpart. Figure \ref{fig:rr_exp} shows learning curves of the RPT models with the RAR method. This figure reveals that RARs always improve the models. In particular, even the RPT1 + RAR model has better learning performance than the RPT3 model. Also, the RPT1 + RAR model and the RPT3 + RAR model use less epochs to be trained, which means that RARs may reduce the time consumption of learning. Figure \ref{fig:crpt_sav_rr_exp} provides learning curves of the models with all improvement methods. A glance at the figure reveals that these models have better performance of learning than the baseline models. Also, they require less epochs to be trained than their counterparts. In addition, the final cross-entropy loss of the C-RPT2 + SAV + RAR model is the lowest among all results.

\begin{table}[!htb]
\caption{Error Rates (\%) After Improvement.}
\label{err_res_2}
\centering
\begin{tabular}{|c|c|c|c|}
\hline
\multirow{2}{*}{Model} & \multicolumn{3}{|c|}{Value of $m$} \\
\cline{2-4}
  & $m=1$ & $m=2$ & $m=3$ \\
\hline
RPT$m$ (Baseline) & 16.5 & 16.6 & 12.9 \\
\hline
RPT$m$ + SAV & 16.8 & 15.8 & 11.5 \\
\hline
RPT$m$ + RAR & 6.7 & 6.0 & 5.4 \\
\hline
RPT$m$ + SAV + RAR & 6.8 & 5.2 & 5.4 \\
\hline
C-RPT$m$ & 16.1 & 12.9 & 11.6 \\
\hline
C-RPT$m$ + SAV & 11.9 & 15.1 & 11.8 \\
\hline
C-RPT$m$ + RAR & 6.3 & 5.1 & 5.1 \\
\hline
C-RPT$m$ + SAV + RAR & 5.4 & $\bm{4.6}$ & 5.3 \\
\hline
\end{tabular}
\end{table}

Table \ref{err_res_2} shows error rates of reasoning on the test set after using the improvement methods. It is noticeable that: Firstly, the C-RPT$m$ models have lower error rates than the baseline RPT$m$ models, especially when $ m = 2 $. Secondly, the RPT$m$ + SAV models have lower error rates than the baseline RPT$m$ model when $ m $ is 2 or 3, but this is not the case for the RPT1 + SAV model. Thirdly, the RARs can reduce the error rates significantly. Finally, the error rates can be reduced further when the three improvement methods are used together. In particular, the C-RPT2 + SAV + RAR model reaches the best error rate (4.6\%) among all models.


\section{Conclusions and Future Work}
\label{conclus}

Deep feedforward neural networks are able to guide rewriting processes after learning from algebraic reasoning schemes. The use of deep structures is necessary, because the behaviours of rewriting can be accurately modelled only if the neural networks have enough hidden layers. Also, it has been shown that the RPT representation is effective for the neural networks to model algebraic expressions, and it can be improved by using the C-RPT representation, the SAV method and the RAR method. Based on these techniques, human-like rewriting can solve many problems about linear equations, differentials and integrals. In the future, we will try to use human-like rewriting to deal with more complex tasks of mathematical reasoning and extend it to more general first-order logics and higher-order logics.

\section*{Acknowledgments}

This work is supported by the Fundamental Research Funds for the Central Universities (No. 2016JX06) and the National Natural Science Foundation of China (No. 61472369).

\bibliographystyle{named}
\bibliography{ijcai17}

\end{document}